\documentclass[letterpaper, 10 pt, conference]{ieeeconf}  % Comment this line out if you need a4paper

\IEEEoverridecommandlockouts                              % This command is only needed if 
\usepackage{graphics} % for pdf, bitmapped graphics files
\usepackage{epsfig} % for postscript graphics files
\usepackage{mathptmx} % assumes new font selection scheme installed
\usepackage{url}
\usepackage{times} % assumes new font selection scheme installed
\usepackage{amsmath} % assumes amsmath package installed
\usepackage{amssymb}  % assumes amsmath package installed
\usepackage[font=footnotesize,labelfont=bf]{caption}
\usepackage[font=footnotesize]{subcaption}
\usepackage{booktabs} % Added
\usepackage{xcolor} % Added
\usepackage{cite} % Added
\usepackage{tabularx}
\usepackage{hyperref}
\newcolumntype{Y}{>{\centering\arraybackslash}X}

\begin{document}
\bstctlcite{IEEEexample:BSTcontrol}

\title{\LARGE \bf
ImagiNav: Scalable Embodied Navigation via Generative Visual Prediction and Inverse Dynamics
% \textcolor{red}{ImagiNav: Geometrically-Grounded Visual Planning for Embodied Navigation}
}

\author{Jie Chen$^{1,2}$, Yuxin Cai$^{2,3}$, Yizhuo Wang$^{1}$, Ruofei Bai$^{2,3}$, Yuhong Cao$^{1}$, Jun Li$^{2}$\\ Wei-Yun Yau$^{2,*}$, and Guillaume Sartoretti$^{1}$%
\thanks{$^{1}$Jie Chen, Yizhuo Wang, Yuhong Cao, and Guillaume Sartoretti are with the Department of Mechanical Engineering, National University of Singapore.}%
\thanks{$^{2}$Jie Chen, Yuxin Cai, Ruofei Bai, Jun Li and Wei-Yun Yau are with the Institute for Infocomm Research (I$^2$R), Agency for Science, Technology and Research (A*STAR), Singapore.}%
\thanks{$^{3}$Yuxin Cai and Ruofei Bai are with Nanyang Technological University.}%
\thanks{$^{*}$Corresponding author.}%
}
% \author{Anonymous IROS Submission}

\maketitle
\thispagestyle{empty}
\pagestyle{empty}

%%%%%%%%%%%%%%%%%%%%%%%%%%%%%%%%%%%%%%%%%%%%%%%%%%%%%%%%%%%%%%%%%%%%%%%%%%%%%%%%
%%%%%%%%%%%%%%%%%%%%%%%%%%%%%%%%%%%%%%%%%%%%%%%%%%%%%%%%%%%%%%%%%%%%%%%%%%%%%%%%
\begin{figure*}[t]
    \vspace{0.15cm}
    \centering
    \includegraphics[width=\textwidth]{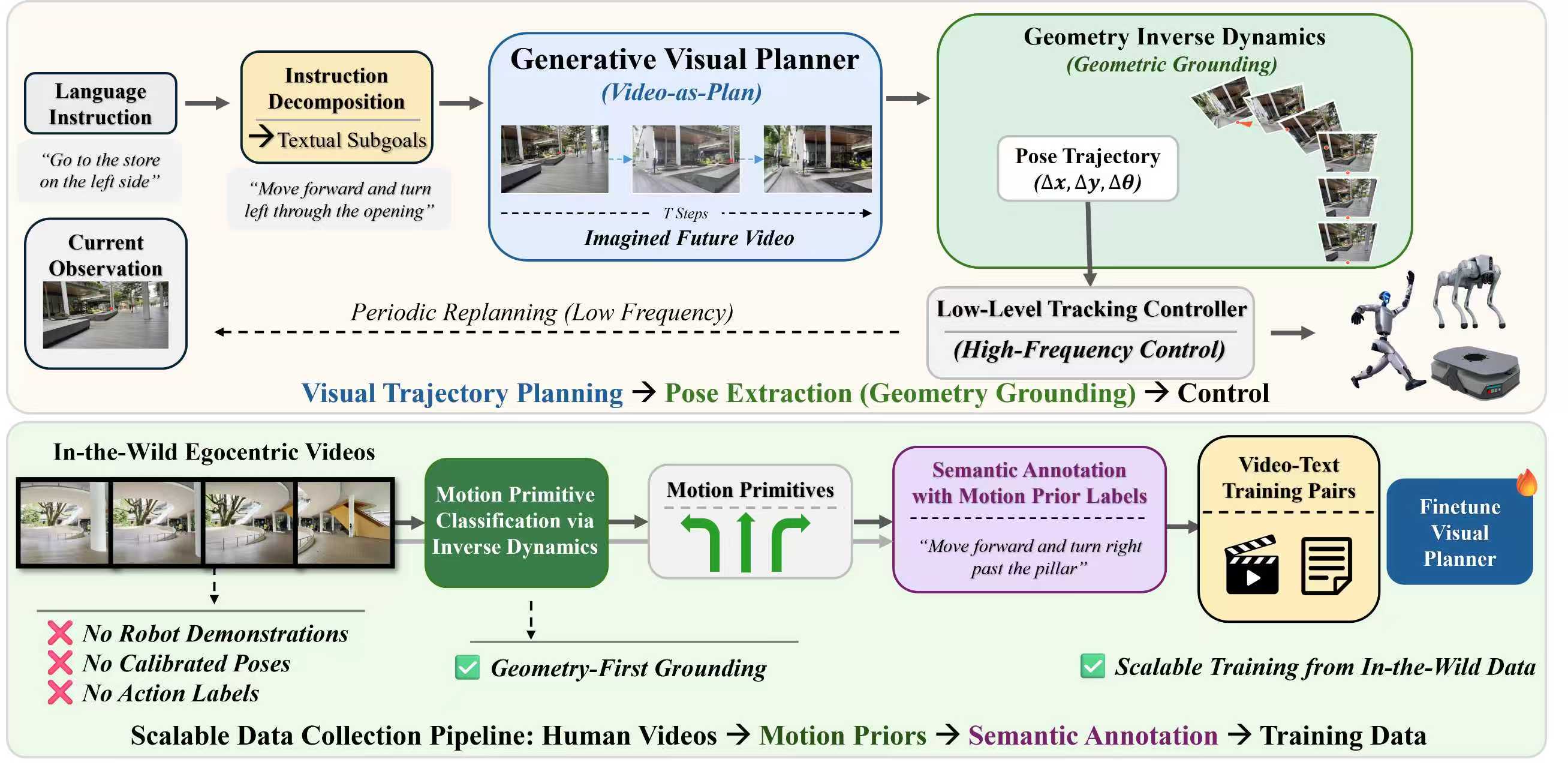}
    % \caption{\textbf{Overview of the ImagiNav Framework.} The system operates hierarchically: (1) A \textbf{System-2 Reasoner} decomposes the high-level instruction into textual subgoals; (2) A \textbf{Generative Video Model} acts as a visual planner, imagining the future video trajectory conditioned on the text; (3) An \textbf{IDM} extracts actionable camera extrinsics from the imagined video, which are then executed by the low-level controller.}
    \caption{\textbf{Overview of the ImagiNav Framework.}
    Top: Navigation is formulated in visual space. A generative visual planner synthesizes future egocentric observations, which are geometrically grounded into pose trajectories via inverse dynamics and executed by a tracking controller.
    Bottom: Geometry-first scalable training pipeline from in-the-wild egocentric videos.
    }
    \vspace{-12pt}
    \label{fig:pipeline}
\end{figure*}

\begin{abstract}

Enabling robots to navigate open-world environments via natural language is critical for general-purpose autonomy. Yet, Vision-Language Navigation has relied on end-to-end policies trained on expensive, embodiment-specific robot data. While recent foundation models trained on vast simulation data show promise, the challenge of scaling and generalizing persists due to the limited scene diversity and visual fidelity in simulation. To address this gap, we propose ImagiNav, a novel hierarchical paradigm that formulates navigation in visual space. Instead of predicting waypoints, ImagiNav synthesizes a future egocentric video conditioned on language instructions, serving as a high-level plan, interpreted by an inverse dynamics model to extract metric trajectories for execution. By decoupling planning from robot actuation, the paradigm enables direct utilization of diverse in-the-wild navigation videos. To support this, we develop an auto-labeling data pipeline that enhances motion annotation accuracy. ImagiNav demonstrates strong zero-shot transfer to robot navigation without requiring robot demonstrations, paving the way for generalist robots that learn navigation directly from unlabeled, open-world data. The project page is available at: \href{https://j1dan.github.io/ImagiNav}{https://j1dan.github.io/ImagiNav}

\end{abstract}

\section{INTRODUCTION}
 
The long-standing goal of Embodied Navigation \cite{anderson2018visionr2r} is to develop general-purpose robots capable of following natural language instructions to navigate diverse environments. Mastering this capability is essential for deploying autonomous agents in complex real-world applications. By relying on natural language for intuitive human-robot interaction and egocentric vision for rich semantic understanding, robots can navigate unstructured spaces without depending on precise metric localization, which is notoriously fragile in dynamic environments. To achieve this, Vision-Language Navigation (VLN) agents must possess two critical capabilities: robust high-level reasoning to interpret instructions and precise low-level control to execute actions. While recent advances in end-to-end learning \cite{wei2025ground} have shown promise, they predominantly rely on policies trained in simulation. Despite significant efforts to develop continuous and physically realistic simulation platforms \cite{krantz2020vlnce, wang2025rethinking}, a fundamental bottleneck persists: the limited scene diversity and visual fidelity inherent to simulation fail to capture the true complexity of the real world, severely restricting deployment in open-world scenarios. Conversely, collecting large-scale real-world robot demonstrations is prohibitively expensive. A vastly more abundant resource to overcome this data bottleneck is in-the-wild egocentric navigation video \cite{liu2025citywalker}. Whether easily collected or sourced from the web, these videos naturally encode rich navigational intents and capture vast environmental diversity, making them an ideal medium for learning open-world navigation. However, utilizing such data for robotic learning presents a unique challenge: videos lack the action labels required to train conventional policies. To address this, we propose a paradigm shift from \textit{action prediction} to \textit{visual imagination}, treating video as the universal interface between reasoning and control. 

In this work, we introduce \textbf{ImagiNav}, a scalable generative visual planning framework that formulates navigation directly in visual space. Departing from traditional action-prediction architectures, ImagiNav synthesizes the future visual observations of the navigation process. This generated video serves as a predictive visual plan, from which an inverse dynamics model (IDM) later extracts the pose trajectories to be tracked by a low-level controller. The key insight is that while robot kinematics vary, the visual outcome of what the agent expects to see while progressing toward a goal is largely embodiment-invariant. By treating videos as plans, we leverage the rich semantic and physical priors inherent in large-scale pretrained video models, while remaining independent of specific control parameterizations. Crucially, this formulation unlocks the vast potential of egocentric navigation videos for scalable robot learning.

To initiate this process, we utilize a pretrained Vision-Language Model (VLM) as a semantic reasoner to decompose the high-level task into textual subgoals. Conditioned on the subgoal and the current egocentric observation, a generative video model synthesizes the visual plan. By finetuning this model on a dataset of handheld smartphone videos captured in everyday environments, we enable the robot to effectively ground the reasoner's abstract intent into physically plausible visual dynamics. To mitigate spatial ambiguity (e.g., confusing ``left'' with ``right''), a common failure mode in video models \cite{luo2025v}, we implement an Action-Conditioned Mixture-of-Experts (AC-MoE) strategy that routes the generation task to specialized experts, yielding precise motion dynamics. To translate imagined visual trajectories into executable actions, ImagiNav incorporates an IDM that serves as a geometric grounding layer. This module decodes relative pose changes directly from the synthesized video into metric waypoints. Operating within a temporal hierarchy, the generative model acts as a low-frequency visual planner, while a standard high-frequency controller simply tracks the decoded waypoints. This visual-to-geometric interface provides an embodiment-agnostic planning-control abstraction while rendering the agent's intent inherently interpretable.
% In practice, the generative model operates as a low-frequency planner, while a high-frequency tracking controller executes the decoded trajectory in closed loop. This visual-to-geometric interface provides an embodiment-agnostic planning-control abstraction.
% }

Building on this representation, we further develop a geometry-first data collection pipeline that enables training from in-the-wild human egocentric navigation videos without requiring precise state estimation. By extracting motion primitives through geometric inverse dynamics prior to semantic annotation, we mitigate spatial hallucination errors common in purely VLM labeling. This paradigm eliminates the need for robot teleoperation, precise localization systems, and metric calibration during data collection, making the approach inherently scalable.
% }
Our results demonstrate that ImagiNav successfully transfers navigation priors learned from in-the-wild human navigation videos to robots. This successful zero-shot transfer, combined with video generation evaluations confirming the superior out-of-distribution generalization of human data over simulation, underscores the potential of massive egocentric navigation videos as a scalable pathway to generalist embodied intelligence that effectively bypasses the bottleneck of expensive, platform-specific robot demonstrations.

\section{RELATED WORK}

\subsection{Vision-Language Navigation}

Vision-Language Navigation (VLN) tasks agents with following natural language instructions in unseen 3D environments \cite{anderson2018visionr2r}. Early works primarily relied on reconstructed indoor scenes \cite{chang2017matterport3d, ramakrishnan2021habitat} or synthetic environments \cite{fu20213d, khanna2024habitat, InteriorAgent2025} within simulators \cite{savva2019habitatplatformembodiedai, krantz2020vlnce}, often simplifying navigation into discrete graph traversal or assuming idealized robot dynamics. Recent advances have incorporated large pretrained language models and transformer-based architectures to enhance high-level reasoning \cite{ cai2025clcotnavclosedloophierarchicalchainofthought, wei2025ground}. While effective in controlled settings, these methods rely heavily on robot data, predominantly from simulation, leading to generalization gaps in diverse real-world environments.

Efforts to reduce domain gaps include more realistic simulation platforms such as VLN-PE \cite{wang2025rethinking, internnav2025} that introduce realistic challenges
like physical shaking. Other approaches focus on leveraging web-scale video data \cite{liu2025citywalker}. However, these approaches typically require either precise state estimation for action labeling or retain action-space supervision, limiting their scalability. In contrast, we reformulate navigation planning in visual space, separating semantic intent from control parameterization.

%%%%%%%%%%%%%%%%%%%%%%%%%%%%%%%%%%%%%%%%%%%%%%%%%%%%%%%%%%%%%%%%%%%%%%%%%%%%%%%%

\subsection{Generative Models}

% tmp removed citation for image and video and vla model {ramesh2022hierarchicaltextconditionalimagegeneration, rombach2022highresolutionimagesynthesislatent} {kong2025hunyuanvideosystematicframeworklarge}, {kim2024openvlaopensourcevisionlanguageactionmodel}

Generative modeling has revolutionized content creation, expanding from static image synthesis \cite{ho2020denoisingdiffusionprobabilisticmodels, saharia2022photorealistictexttoimagediffusionmodels} and dynamic video generation \cite{hacohen2024ltxvideorealtimevideolatent, wan2025wanopenadvancedlargescale} to complex robotic planning. In the domain of robotics, diffusion models effectively represent multi-modal action distributions \cite{chi2024diffusionpolicyvisuomotorpolicy, cai2025navdplearningsimtorealnavigation}, while Vision-Language-Action models \cite{intelligence2025pi05visionlanguageactionmodelopenworld} advance this by leveraging large backbones to inherit rich semantic priors for action diffusion. Recent benchmark studies have also evaluated the planning capabilities of video models \cite{yue2025ewmbench, wang2025target}.

Beyond direct policy learning, video generation has emerged as a world model, capable of simulating interactive environments \cite{yang2023learning} or serving as a trajectory scorer for navigation agents \cite{bar2025navigationworldmodels}. Distinct from methods that use these models primarily for simulation, we directly leverage video generation as a policy, functioning as a language-conditioned world model that imagines the future trajectory.

\subsection{Inverse Dynamics from Visual Observations}

% Visual Odometry (VO) \cite{Mur_Artal_2017, Qin_2018, Campos_2021} and Structure-from-Motion (SfM) \cite{Schonberger2016SfM} have traditionally served as the backbone of state estimation. In the context of learning from observation, VO can be effectively re-framed as an IDM: rather than merely localizing an agent, the estimated relative pose change serves as a proxy for the high-level action required to transition between states.

% Recent advances in geometric foundation models \cite{keetha2026mapanythinguniversalfeedforwardmetric, wang2025vggtvisualgeometrygrounded}, have scaled this capability by training large-scale Transformers on diverse video corpora. These architectures unify pose estimation and dense reconstruction into a single feed-forward pass, demonstrating strong zero-shot generalization. We leverage this class of foundation models as off-the-shelf IDMs, decoding the "imagined" video trajectory into executable velocity commands to bypass embodiment-specific action labeling.

% \textcolor{red}{
Visual Odometry (VO) and Structure-from-Motion (SfM) techniques have long been used for estimating camera motion from image sequences \cite{Campos_2021, Schonberger2016SfM}. Recent geometric foundation models further scale pose estimation by leveraging large video corpora and transformer architectures \cite{wang2025vggtvisualgeometrygrounded}, enabling robust relative pose prediction in diverse environments. Rather than using geometric estimation purely for localization, we reinterpret inverse dynamics as a geometric grounding layer that translates visual trajectories into physically executable motion commands. This design enables us to bridge generative visual planning with embodiment-agnostic control without requiring explicit action supervision.

% tmp removed citation for SLAM {Mur_Artal_2017, Qin_2018}

%%%%%%%%%%%%%%%%%%%%%%%%%%%%%%%%%%%%%%%%%%%%%%%%%%%%%%%%%%%%%%%%%%%%%%%%%%%%%%%%
%%%%%%%%%%%%%%%%%%%%%%%%%%%%%%%%%%%%%%%%%%%%%%%%%%%%%%%%%%%%%%%%%%%%%%%%%%%%%%%%

\section{IMAGINAV}

The ImagiNav framework is structured as a hierarchical planning architecture comprising three interconnected modules: Semantic Reasoning, Visual Imagination, and Geometry Inverse Dynamics. The overall architecture is illustrated in Figure~\ref{fig:pipeline}.

% \textcolor{red}{
% ImagiNav formulates embodied navigation as planning in \emph{visual trajectory space}, followed by geometric grounding into executable motion. 
% Instead of directly predicting embodiment-specific actions or metric waypoints, the framework synthesizes a future egocentric visual trajectory and subsequently decodes it into relative pose displacements via inverse dynamics. 
% As illustrated in Figure~\ref{fig:pipeline}, the system consists of three conceptual stages: 
% (1) visual trajectory planning, 
% (2) geometric grounding through inverse dynamics, and 
% (3) low-level tracking control. 
% An optional semantic prompting module may decompose long-horizon instructions into short-horizon textual descriptions, but the core contribution lies in the visual-to-geometric planning interface.
% }

%%%%%%%%%%%%%%%%%%%%%%%%%%%%%%%%%%%%%%%%%%%%%%%%%%%%%%%%%%%%%%%%%%%%%%%%%%%%%%%%

\subsection{Semantic Reasoning}

We leverage a pretrained VLM to act as the hierarchical task decomposer. Operating on the current egocentric frame and the global navigation instruction, this module formulates the navigational intent into a natural language prompt. This textual output acts as the semantic condition for the video generation module, translating high-level reasoning into a specific description of the expected future visual trajectory.

% \subsection{Semantic Prompting}
% \textcolor{red}{
% To stabilize long-horizon instruction following, we employ a pretrained vision-language model to transform global navigation instructions into short-horizon textual prompts. 
% Given the current observation $I_t$ and instruction $T_{global}$, the module outputs a concise description of the immediate navigational intent, denoted as $T_{sub}$. 
% }
% \textcolor{red}{
% This text serves as conditioning input to the visual trajectory planner. 
% Importantly, this component is implementation-agnostic and can be replaced by prompt templates or alternative language models. 
% Our framework does not rely on any specific reasoning architecture; the planning mechanism operates entirely in visual trajectory space.
% }

%%%%%%%%%%%%%%%%%%%%%%%%%%%%%%%%%%%%%%%%%%%%%%%%%%%%%%%%%%%%%%%%%%%%%%%%%%%%%%%%

\subsection{Visual Imagination}

The core planning component is a generative video model $\mathcal{G}$ that operates in visual space. Given the current observation $I_t$ and a textual instruction $L_t$, the model predicts a sequence of future egocentric frames:
\begin{equation}
    V_{pred} = \mathcal{G}(I_t, L_t),
    \label{eq:generation}
\end{equation}
% where $I_t$ represents the feature of the current frame and $T_{sub}$ provides the textual condition.
% \textcolor{blue}{
where $V_{pred} = \{I_{t+1}, \dots, I_{t+H}\}$ represents an imagined future visual trajectory over a horizon of $H$ steps. 
Unlike conventional policies that output control signals or waypoints, $V_{pred}$ encodes the navigation plan implicitly through expected visual progression.

Specifically, we instantiate $\mathcal{G}$ using the pre-trained LTX-2B architecture, which employs a Diffusion Transformer backbone \cite{peebles2023scalable} operating within a compressed spatiotemporal latent space. The generation process follows a rectified flow matching objective \cite{lipman2023flowmatchinggenerativemodeling, liu2022flowstraightfastlearning}, where a latent representation $z_{\tau}$ is evolved from a noise prior $z_1 \sim \mathcal{N}(0, \mathbf{I})$ to a clean latent $z_0$:
% }
\begin{equation}
    \frac{d z_{\tau}}{d \tau} = v_{\theta}(z_{\tau}, \tau, c),
\end{equation}
where $v_{\theta}$ denotes the transformer-based velocity field prediction parameterized by $\theta$, conditioned on the encoded start frame and text instruction $c = \{\mathcal{E}(I_t), \mathcal{E}(L_t)\}$. The final video is reconstructed via a causal Video-VAE decoder that performs the final denoising step during latent-to-pixel decoding. To ensure domain relevance, we finetune $\theta$ on our scalable, human-collected video dataset (detailed in Section~\ref{sec:data}), aligning the generative flow with physically plausible ego-motion dynamics. To implement the AC-MoE strategy for mitigating the spatial ambiguity common in video generation, we bifurcate our training data to finetune two specialized expert models. Both experts handle forward motions but specialize in left and right turns, respectively. During inference, the semantic reasoner dynamically routes the generation task to the appropriate expert based on the intended subgoal.

The generative model grounds high-level textual subgoals into synthesized visual trajectories, explicitly encoding the robot's navigational intent and affording highly modular integration into downstream control pipelines. For instance, key frames could be extracted to serve as target observations for a learned image-goal policy. In our proposed architecture, we bypass intermediate visual policies by directly decoding the trajectory from the synthesized video via an IDM.

%%%%%%%%%%%%%%%%%%%%%%%%%%%%%%%%%%%%%%%%%%%%%%%%%%%%%%%%%%%%%%%%%%%%%%%%%%%%%%%%

\subsection{Geometry Inverse Dynamics}

To bridge visual planning and physical execution, we introduce a geometric grounding layer that extracts an ego-motion trajectory from the video plan. Specifically, an IDM $\Phi$ decodes the synthesized video $V_{pred}$ into a sequence of relative waypoints:
\vspace{-0.1cm}
\begin{equation}
\mathbf{T}_{plan} = \{(\Delta \tilde{x}_i, \Delta \tilde{y}_i, \Delta \theta_i)\}_{i=1}^N = \Phi(V_{pred}).
\label{eq}
\end{equation}
Here, $(\Delta \tilde{x}_i, \Delta \tilde{y}_i, \Delta \theta_i)$ represents the scale-normalized relative ego-motion at step $i$. The resulting trajectory provides an embodiment-agnostic navigation interface, which is subsequently converted into executable robot commands by a low-level controller with appropriate scale calibration and kinematic constraints.

% \subsection{Low-Level Tracking Control}
% \textcolor{blue}{
% The extracted pose trajectory is executed by a high-frequency tracking controller that converts relative displacements into motor commands compatible with the robot's hardware interface. 
% While visual planning operates at a low frequency over horizon $H$, control is executed in closed loop to compensate for actuation noise and environmental perturbations.
% At runtime, the system performs periodic replanning: after executing a segment of the decoded trajectory, the updated observation is fed back into the visual planner to generate a new imagined trajectory. 
% This receding-horizon strategy balances computational efficiency with robustness to dynamic changes.
% }

%%%%%%%%%%%%%%%%%%%%%%%%%%%%%%%%%%%%%%%%%%%%%%%%%%%%%%%%%%%%%%%%%%%%%%%%%%%%%%%%
%%%%%%%%%%%%%%%%%%%%%%%%%%%%%%%%%%%%%%%%%%%%%%%%%%%%%%%%%%%%%%%%%%%%%%%%%%%%%%%%

\section{SCALABLE DATA COLLECTION PIPELINE}
\label{sec:data}

% A fundamental bottleneck to generalizable VLN is the scarcity of diverse, real-world robotic data. While simulation provides scale, it suffers from a significant sim-to-real visual gap and limited scene diversity (often restricted to residential environments). Conversely, collecting real-world robot demonstrations via teleoperation is prohibitively expensive and difficult to scale.
% \textcolor{red}{
A key advantage of planning in visual space is that supervision no longer requires embodiment-specific action labels or pose annotations. 
Because the planner operates purely in visual space, learning reduces to modeling physically plausible visual progression conditioned on text.
This observation enables a shift from metric-annotated data collection to egocentric video curation.
% }

% By employing a video generation model as the primary planner, we circumvent the strict requirement for ground-truth camera extrinsics, which are difficult to obtain in the wild. This architectural choice enables a paradigm shift from \textit{robot-centric} collection to \textit{human-centric} curation. We leverage the ubiquity of consumer-grade cameras (e.g., smartphones, GoPros) to compile a massive dataset of real-world human walking trajectories. This approach captures high-fidelity visual dynamics and diverse environmental layouts without requiring robotic hardware.
% \textcolor{red}{
Instead of collecting expensive robot demonstrations, we leverage in-the-wild human walking videos captured using commodity cameras (e.g., smartphones and GoPros). 
These videos naturally contain rich ego-motion dynamics and diverse environmental layouts, without requiring robotic hardware, localization systems, or camera calibration.
% }

%%%%%%%%%%%%%%%%%%%%%%%%%%%%%%%%%%%%%%%%%%%%%%%%%%%%%%%%%%%%%%%%%%%%%%%%%%%%%%%%

\subsection{Auto-Labeling via IDM-Guided Annotation}
\label{subsec:autolabel}
% A critical challenge in utilizing raw human videos is the absence of ground-truth action labels and textual instructions. While recent works employ VLMs to auto-label video data, we identify a significant failure mode: VLMs frequently struggle with precise spatial reasoning, often hallucinating motion directions (e.g., misinterpreting a left turn as a right turn).

% To mitigate this, we propose a \textit{Geometry-First, Semantics-Second} labeling pipeline. We treat the IDM as a geometric verifier to enforce kinematic accuracy, while leveraging the LLM solely for semantic contextualization. The pipeline proceeds in two stages:

% \textcolor{blue}{
Raw human videos lack textual navigation instructions. 
Directly relying on VLMs for annotation introduces spatial hallucination errors, particularly in directional reasoning (e.g., confusing left and right turns). To address this, we adopt a \textit{geometry-first, semantics-second} supervision pipeline. 
Geometric motion primitives are extracted prior to semantic labeling, ensuring that textual descriptions remain physically grounded.
% }

\subsubsection{Geometric Motion Primitive Extraction}
% We first pass the raw video frames through a pre-trained IDM to extract the camera's ego-motion. Based on the extracted trajectory, we employ a heuristic classifier to categorize the clip into one of 7 discrete motion primitives: 
% \textit{Move Forward}, \textit{Turn Left/Right}, \textit{Forward + Turn Left/Right}, and \textit{Turn Left/Right + Forward}. 
% This step ensures the motion label is grounded in physical geometry rather than visual interpretation.
% \textcolor{red}{
We first segment raw videos into fixed-length clips and apply a pretrained IDM to recover relative ego-motion trajectories. Based on these poses, a heuristic classifier then categorizes the motion patterns into distinct primitives, including forward translation, rotational turns, and compound movements. This step ensures motion labels are grounded in physical geometry rather than visual interpretation. Importantly, this step does not require explicit camera calibration or robot-specific kinematic modeling. All motion signals are derived directly from image-space geometry.

\subsubsection{Motion-Primitive–Guided Captioning}

% Leveraging the extracted motion primitive as auxiliary information, we prompt Gemini to generate a dense, natural language description of the scene (e.g., \textit{``Turn left and move forward down the corridor''}). Crucially, to better align with the cinematographic data distribution seen during the video model's pretraining, we map generic navigation terms to camera movement terminology, replacing \textit{``move''} with \textit{``dolly''} and \textit{``turn''} with \textit{``pan''}.
% \textcolor{blue}{
Given the extracted motion primitives, we prompt a pretrained large language model to generate dense scene descriptions that are consistent with the recovered motion. 
The geometric motion prior constrains the semantic annotation process, mitigating spatial hallucination errors. To better align with the distribution of video model pretraining data, we map navigation verbs to camera movement terminology (e.g., replacing \textit{``move''} with \textit{``dolly''} and \textit{``turn''} with \textit{``pan''}). This results in a scalable, low-cost dataset $\mathcal{D} = \{(V_{clip}, T_{desc})\}$ where the visual trajectory is physically consistent with the text, enabling the video model to learn robust navigation dynamics without direct robot supervision.

%%%%%%%%%%%%%%%%%%%%%%%%%%%%%%%%%%%%%%%%%%%%%%%%%%%%%%%%%%%%%%%%%%%%%%%%%%%%%%%%
%%%%%%%%%%%%%%%%%%%%%%%%%%%%%%%%%%%%%%%%%%%%%%%%%%%%%%%%%%%%%%%%%%%%%%%%%%%%%%%%

\section{EXPERIMENTS}
\label{sec:exp}

% We evaluate the proposed ImagiNav framework on both high-fidelity simulation and investigate how different sources of data could affect the video model, its core component's performance by comparing visual quality and the motion fidelity between real-world collected trajectories of the generated videos.
% \textcolor{red}{
We evaluate ImagiNav from two complementary perspectives. First, we assess whether the visual planner can leverage its learned priors to achieve zero-shot generalization from real-world human videos to unseen simulated environments, driving a physically simulated robot in a high-fidelity embodied benchmark. Second, we study how the source of finetuning data (simulation vs.\ human videos) affects the video planner's visual quality and kinematic realism.
% }

%%%%%%%%%%%%%%%%%%%%%%%%%%%%%%%%%%%%%%%%%%%%%%%%%%%%%%%%%%%%%%%%%%%%%%%%%%%%%%%%

\subsection{Simulation Experimental Setup}
% We evaluate the proposed ImagiNav framework within the VLN-PE benchmark, a platform built on NVIDIA Isaac Sim offering high-fidelity visual rendering and physically realistic dynamics. All methodologies are evaluated using a simulated Unitree H1 humanoid robot embodiment. The low-level locomotion controller is trained via reinforcement learning, accepting linear and angular velocity commands to execute robust bipedal locomotion. We evaluate the methods on 100 episodes from the InteriorNav dataset built on the InteriorAgent scene that offers high visual fidelity.
% \textcolor{red}{
We evaluate ImagiNav in the VLN-PE benchmark \cite{wang2025rethinking, internnav2025}, built on NVIDIA Isaac Sim \cite{liang2018gpuacceleratedroboticsimulationdistributed}, providing high-fidelity rendering and physically realistic dynamics.
All methods are evaluated on the same simulated Unitree H1 humanoid embodiment.
The low-level locomotion policy is trained via reinforcement learning and takes linear and angular velocity commands to execute bipedal walking.
We report VLN-PE standard navigation and stability metrics on 100 episodes from the InteriorNav dataset, which utilizes the high-quality InteriorAgent scenes \cite{InteriorAgent2025}.
% }

%%%%%%%%%%%%%%%%%%%%%%%%%%%%%%%%%%%%%%%%%%%%%%%%%%%%%%%%%%%%%%%%%%%%%%%%%%%%%%%%

\subsection{Implementation Details}
\label{sec:implementation}

\begin{figure*}[t] % Added the asterisk to span both columns
    \centering
    % Left Plot: Now 48% of the full page width
    \begin{subfigure}[b]{0.38\textwidth}
        \centering
        \includegraphics[width=\linewidth]{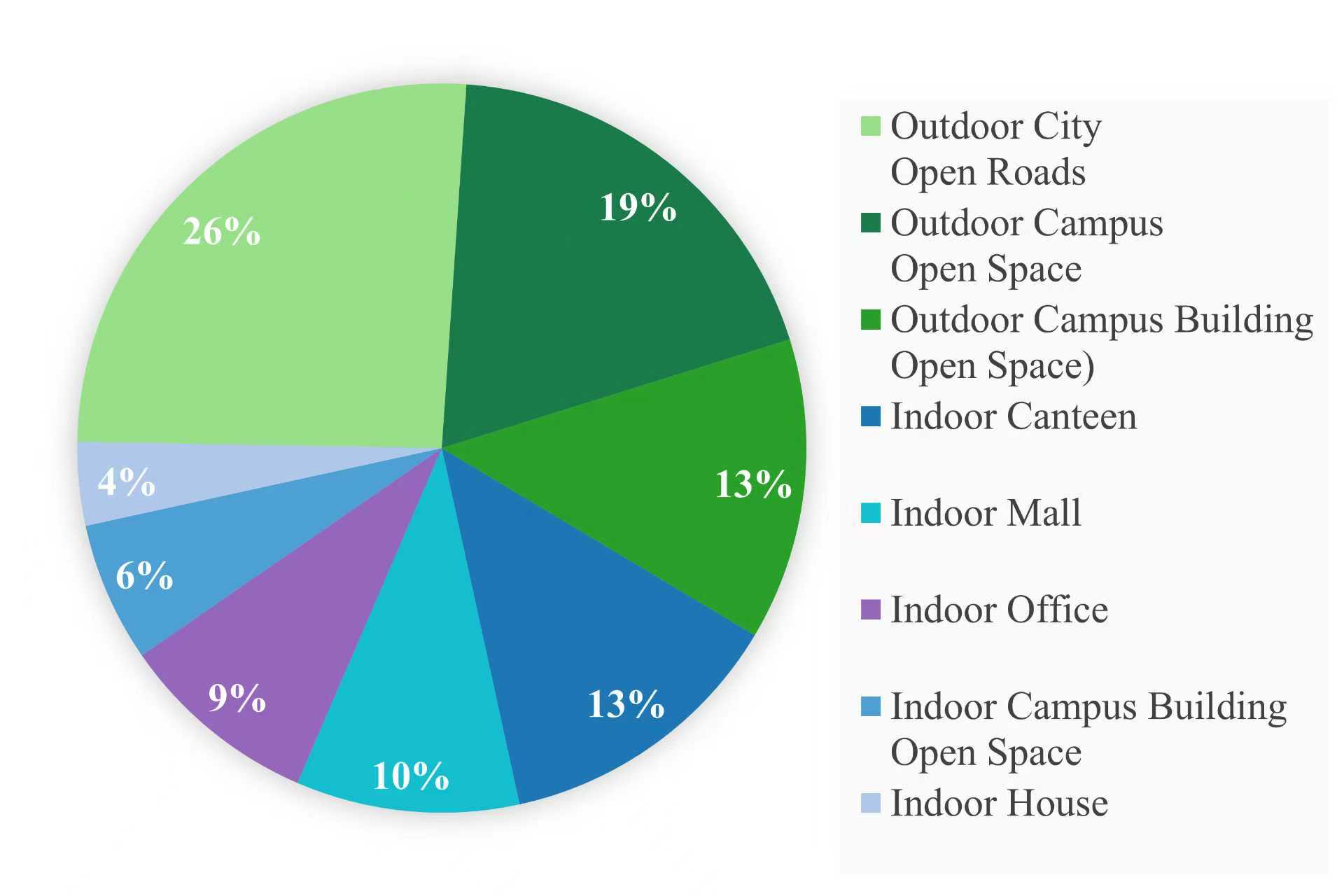}
        \caption{Scene Distribution}
        \label{fig:data_dist_scene}
    \end{subfigure}
    \hfill
    % Right Plot: Now 48% of the full page width
    \begin{subfigure}[b]{0.38\textwidth}
        \centering
        \includegraphics[width=\linewidth]{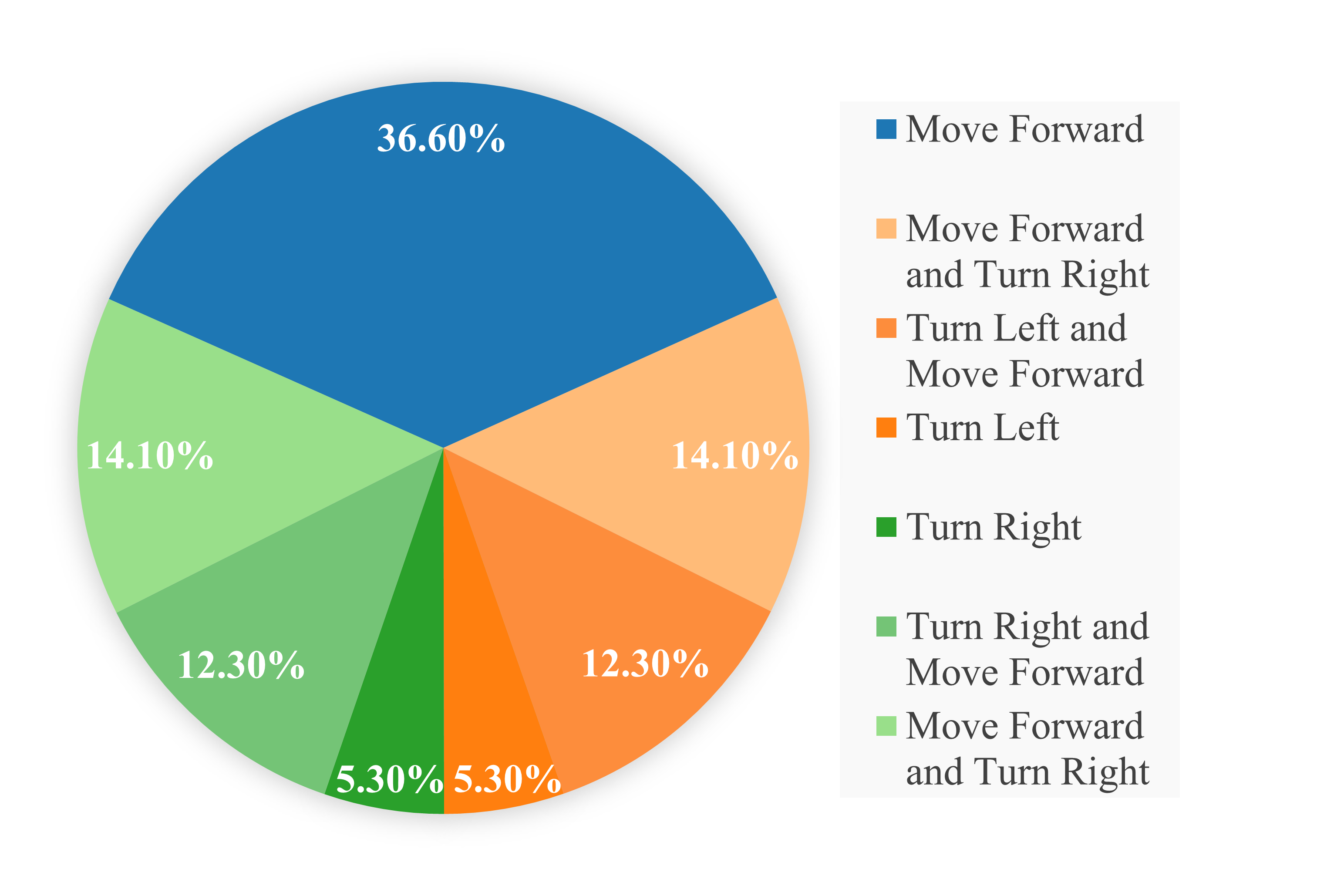}
        \caption{Motion Primitive Distribution}
        \label{fig:data_dist_motion}
    \end{subfigure}
    
    \caption{\textbf{Dataset Statistics.} \textbf{(a)} Distribution of raw videos across different real-world environments. \textbf{(b)} Distribution of motion primitives in the final augmented dataset, showing balanced coverage.}
    \label{fig:dataset_stats}
    \vspace{-0.5cm}
\end{figure*}

% \noindent \textbf{Data Collection and Augmentation.} We collect training data using a hand-held iPhone, capturing casual navigation sequences in diverse environments such as shopping malls and university campuses, totaling around 33 minutes. The video clips used to finetune the video model are sampled from the raw videos and labeled via the method mentioned in~\ref{subsec:autolabel}. We curated a diverse dataset spanning various real-world environments and motion primitives. The distribution of collected scenes, categorized by location type, is visualized in Figure~\ref{fig:data_dist_scene}. As shown, the dataset covers a wide range of indoor and outdoor settings, from university campuses to commercial spaces. Furthermore, to mitigate class imbalance in the raw data, we applied geometric augmentation (mirroring). The resulting distribution of motion primitives in the augmented dataset is presented in Figure~\ref{fig:data_dist_motion}, demonstrating a balanced coverage of navigational actions.
% \textcolor{blue}{
\noindent \textbf{Real-World Data Collection.}
To validate the proposed pipeline, we collect a pilot-scale dataset using a hand-held iPhone in diverse environments (e.g., shopping malls and university campuses).
Video clips are sampled from the raw videos of approximately 33 minutes, and labeled using the IDM-guided procedure in Section~\ref{subsec:autolabel}.
To reduce class imbalance, we apply geometric mirroring augmentation, yielding a more balanced coverage of motion primitives (Fig.~\ref{fig:dataset_stats}), resulting in 2,558 clips of 121 frames each.
% }

\noindent \textbf{System Instantiation.}
We instantiate the abstract modules of ImagiNav with foundation models:
\begin{itemize}
    \item \textbf{Reasoner:} We use Gemini 3.0 Flash \cite{google2025gemini3} for high-level task decomposition.
    \item \textbf{Video Generation Model ($\mathcal{G}$):} We employ the LTX-Video-2B architecture \cite{hacohen2024ltxvideorealtimevideolatent}, a latent diffusion model parameterized by a Diffusion Transformer backbone. To adapt the model to the domain of ego-centric navigation, we finetune the pretrained model on our collected dataset using Low-Rank Adaptation (LoRA) \cite{hu2022lora}. The model is configured to synthesize video sequences with a spatial resolution of $480 \times 256$ (width $\times$ height) and a temporal depth of 121 frames, operating at a frame rate of 24 Hz.
    \item \textbf{Inverse Dynamics Model ($\Phi$):} We employ VGGT \cite{wang2025vggtvisualgeometrygrounded}, a geometric foundation model, to extract relative camera poses from the generated videos, where the frames are sampled at an interval. 
\end{itemize}
During inference, we utilize a 70-frame functional video generation horizon, where the IDM operates on frames subsampled at a stride of 4.

\noindent \textbf{Low-Level Controller.}
While ImagiNav is compatible with various path-tracking algorithms, we employ a simple heuristic proportional controller for the experiments. To map the IDM-predicted displacements $(\Delta x, \Delta y, \Delta \theta)$ to the robot's velocity command $(v, \omega)$, we use:
\begin{equation}
   v =
   \begin{cases}
    \text{clip}(k_v \sqrt{\Delta x^2 + \Delta y^2}, 0, v_{max}) & \text{if } |\Delta \theta| \leq \frac{\pi}{4} \\
    0 & \text{if } |\Delta \theta| > \frac{\pi}{4}
   \end{cases},
\end{equation}
\begin{equation}
   \omega = \text{clip}(k_{\theta} \cdot \text{wrap}(\Delta \theta), -\omega_{max}, \omega_{max}),
\end{equation}
where $k_v$ and $k_{\theta}$ are proportional gains.

%%%%%%%%%%%%%%%%%%%%%%%%%%%%%%%%%%%%%%%%%%%%%%%%%%%%%%%%%%%%%%%%%%%%%%%%%%%%%%%%

\subsection{Baselines}

To rigorously assess ImagiNav's performance, we benchmark it against zero-shot modular baselines, state-of-the-art domain-trained policies, and a simulation-trained variant of our own method to isolate data efficacy.

\begin{itemize}
    \item \textbf{NavDP \cite{cai2025navdplearningsimtorealnavigation}:} We compare against NavDP, a state-of-the-art diffusion-based policy trained on massive simulation data. It processes RGB-D observations alongside a goal coordinate $(x_{goal}, y_{goal})$ to output a trajectory of future waypoints $\mathbf{T} = \{(\Delta x_i, \Delta y_i, \Delta \theta_i)\}_{i=1}^H$:
    \begin{equation}
    \mathbf{T} = \pi_{\text{NavDP}}(I_{\text{RGB}}, I_{\text{D}}, x_{\text{goal}}, y_{\text{goal}}).
    \end{equation}

    \item \textbf{InternVLA-N1:} We evaluate InternVLA-N1, the first open dual-system navigation foundation model. It employs a finetuned 7B VLM as a system-2 planner and a diffusion-based navigation policy as the system-1 local planner. Trained on the massive InternData-N1 dataset, it achieves state-of-the-art performance on various benchmarks.

    \item \textbf{ImagiNav-Sim:} To investigate the role of real-world video data versus simulation data, we finetune this variant exclusively on simulation data from a subset of the InternData-N1 dataset. We carefully curate samples from \textit{3D-Front}, \textit{R2R}, and \textit{HSSD} scenes to strictly match the total size and motion primitive distribution of the real-world dataset.
\end{itemize}

For all zero-shot evaluations, we adopt an open-loop local execution strategy: the agent synthesizes a trajectory based on the current observation and executes the derived sequence of motor commands. The robot acquires a fresh observation and initiates re-planning after the current action buffer is fully consumed.

%%%%%%%%%%%%%%%%%%%%%%%%%%%%%%%%%%%%%%%%%%%%%%%%%%%%%%%%%%%%%%%%%%%%%%%%%%%%%%%%

\subsection{Simulation Results}

We employ a suite of metrics to evaluate navigation performance. We report \textbf{Success Rate (SR)} for stops within a goal threshold, and Success weighted by Path Length (\textbf{SPL}), measuring efficiency relative to the optimal path, defined as $SPL = \frac{1}{N} \sum_{i=1}^N S_i \frac{\ell_i}{\max(p_i, \ell_i)}$, where $S_i$ is a binary success indicator, $\ell_i$ is the shortest path distance, and $p_i$ is the agent's path length. We also report \textbf{Oracle Success Rate (OS)}, which assesses if any point on the path fell within that threshold, \textbf{Navigation Error (NE)} (distance to goal in meters), and \textbf{Trajectory Length (TL)}.

% We employ a comprehensive suite of metrics to evaluate navigational accuracy and physical stability:
% \begin{itemize}
%    \item \textbf{Navigational Accuracy:} We report \textbf{Success Rate (SR)}, indicating how often the agent stops within a predefined threshold of the goal, and \textbf{Oracle Success Rate (OS)}, which assesses if any point on the path fell within that threshold. We also measure \textbf{Navigation Error (NE)} (distance to goal in meters) and \textbf{Trajectory Length (TL)}.
%    \item \textbf{Physical Stability:} To assess the agent's embodied control, we report \textbf{Fall Rate (FR)}, measuring the frequency of falls, and \textbf{Stuck Rate (StR)}, quantifying instances where the agent becomes immobilized.
% \end{itemize}

\begin{table*}[t]
\centering
\vspace{0.2cm}
\caption{Quantitative Results on InteriorNav Dataset. We report performance relative to input modality (Depth), domain alignment (Sim$\to$Sim), and data scale.}
\label{tab:internnav_quant}
\renewcommand{\arraystretch}{1.2}
\begin{tabularx}{\textwidth}{l cc c YYYYY} % Y columns share equal remaining space
\toprule
\textbf{Method} & \textbf{Depth} & \textbf{Sim $\to$ Sim} & \textbf{Data Size} & \textbf{TL} & \textbf{NE} $\downarrow$ & \textbf{OS} $\uparrow$ & \textbf{SR} $\uparrow$ & \textbf{SPL} $\uparrow$ \\
\midrule
\multicolumn{9}{l}{\textit{In-Domain Supervised}} \\
\midrule
InternVLA-N1 \cite{wei2025ground} & \checkmark & \checkmark & $\sim$1000h & 6.42 & \textbf{3.48} & \textbf{0.63} & \textbf{0.53} & \textbf{0.44} \\
\midrule
\multicolumn{9}{l}{\textit{Zero-Shot Transfer}} \\
\midrule
NavDP \cite{cai2025navdplearningsimtorealnavigation} & \checkmark & \checkmark & $\sim$101h & 3.48 & 4.19 & 0.37 & 0.32 & 0.31 \\
ImagiNav-Sim & $\times$ & \checkmark & $\sim$0.5h & 3.92 & \textbf{3.94} & \textbf{0.45} & \textbf{0.39} & \textbf{0.37} \\
ImagiNav-Real & $\times$ & $\times$ & $\sim$0.5h & 4.03 & 4.13 & 0.41 & 0.36 & 0.35 \\
\bottomrule
\multicolumn{9}{l}{\footnotesize \textit{Note:} TL: Trajectory Length, NE: Navigation Error (m), OS: Oracle Success Rate, SR: Success Rate, SPL: Success weighted by Path Length.} \\
\multicolumn{9}{l}{\footnotesize Best results for each category are \textbf{bolded}.}
\end{tabularx}
\vspace{-0.2cm}
\end{table*}

Our quantitative results are presented in Table~\ref{tab:internnav_quant}. As expected, InternVLA-N1, benefiting from massive in-domain supervision ($\sim$1000 hours), establishes the performance upper bound. In the zero-shot transfer regime, the results highlight distinct performance patterns:

\textbf{1. Modality Gap in Modular Baselines.} Although NavDP is a state-of-the-art visual navigation policy, it slightly underperforms ImagiNav in the VLN setting where the point goal is given by the semantic reasoner. We attribute this performance gap to the inherent deficiency of VLMs in grounding semantic understanding into precise metric coordinates. In the NavDP pipeline, the VLM is tasked with abstracting semantic subgoals into geometric coordinates, which directly exposes the language model's lack of fine-grained spatial reasoning capabilities. In contrast, ImagiNav bridges this gap via visual imagination: by generating the future video explicitly, it grounds the VLM's textual reasoning into a dense, pixel-perfect trajectory.

\begin{figure}[t]
    \centering
    \includegraphics[width=\linewidth]{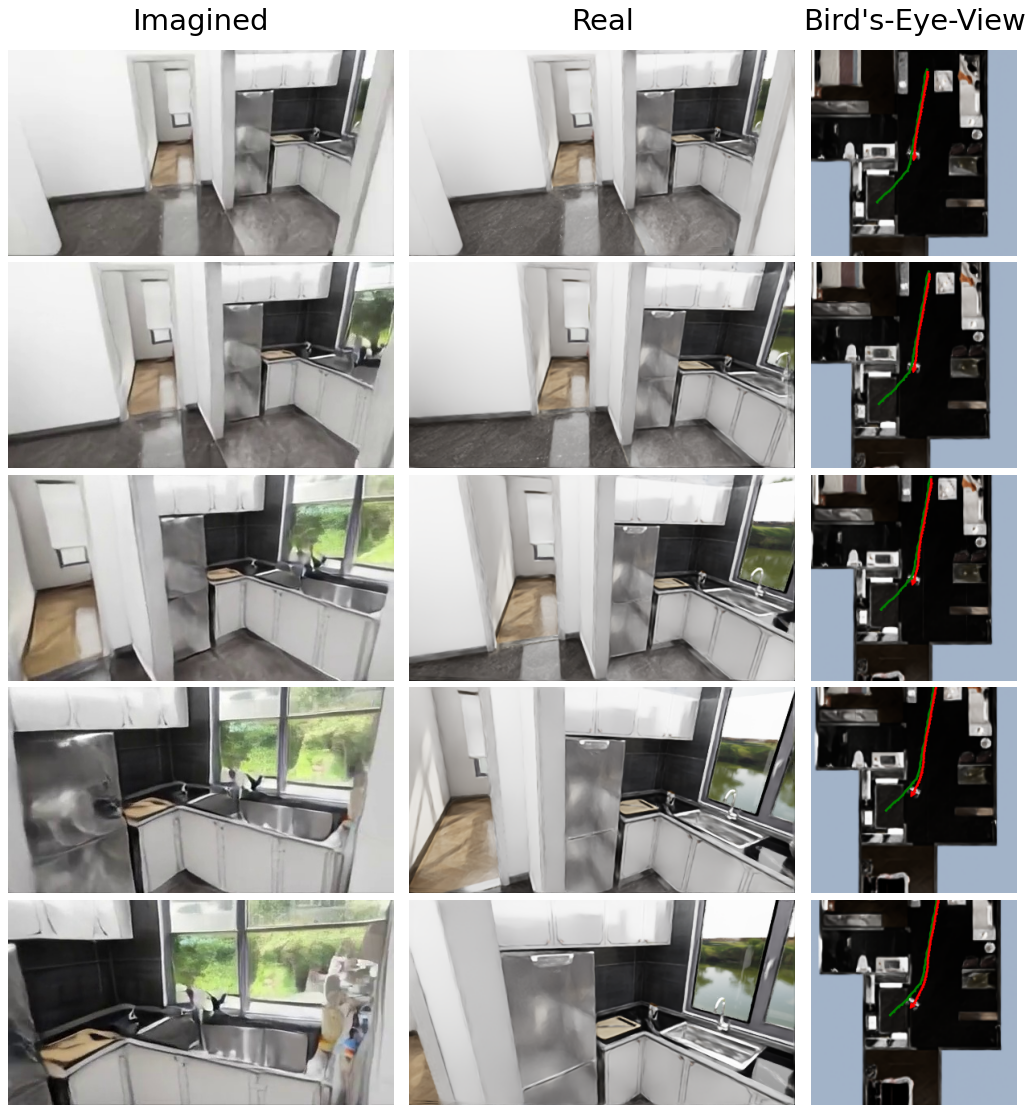}
    \caption{\textbf{Qualitative Visualization of the Imagination-Action Cycle.} 
    \textbf{(Left)} The \textit{imagined} future frame generated by the model. 
    \textbf{(Middle)} The actual robot observation after executing the extracted plan. 
    \textbf{(Right)} Top-down trajectory map, where the \textcolor{red}{red} line indicates the executed path and the \textcolor{green}{green} line represents the reference ground truth. The close alignment confirms the physical consistency of the generated plans.}
    \label{fig:imagined_real}
    \vspace{-10pt}
\end{figure}

\textbf{2. Strong Zero-Shot Generalization.} 
Despite being trained on out-of-domain human videos with negligible exposure to residential indoor scenes, \textbf{ImagiNav-Real} demonstrates robust transferability to the VLN-PE benchmark. This confirms that our planner leverages strong semantic priors from the pretrained video foundation model to generalize the concept of navigable space without overfitting to specific textures. In particular, Figure~\ref{fig:imagined_real} presents a qualitative visualization of the imagination-action cycle. The high visual correspondence between the imagined and actual frames demonstrates the model's ability to synthesize physically feasible plans that are realizable by the low-level controller.

As expected, \textbf{ImagiNav-Sim}, benefiting from direct exposure to in-domain indoor layouts, slightly outperforms ImagiNav-Real. However, the performance gap is remarkably narrow. We attribute this to the kinematic artifacts inherent to simulation datasets. Trajectories generated by deterministic planners like A* on discrete graphs often exhibit unnatural, oscillatory zig-zag patterns and instantaneous directional changes, while our real-world dataset possesses natural, continuous momentum. This 
suggests that, while in-domain simulation data provides better visual layout priors, real-world human data offers superior motion dynamics, partially offsetting this domain mismatch.

%%%%%%%%%%%%%%%%%%%%%%%%%%%%%%%%%%%%%%%%%%%%%%%%%%%%%%%%%%%%%%%%%%%%%%%%%%%%%%%%
\begin{table*}[t]
\vspace{0.2cm}
\centering
\caption{Quantitative Evaluation of LTX-2B Video Model with Sim/Real Data}
\label{tab:video_metrics}
% Increase horizontal padding for better separation
\setlength{\tabcolsep}{7pt}
% Increase vertical padding for readability
\renewcommand{\arraystretch}{1.2}
\begin{tabular}{l|c c c c c c c}
\hline
\textbf{Model Variant} & \textbf{FVD} $\downarrow$ & \textbf{LPIPS} $\downarrow$ & \textbf{PSNR} $\uparrow$ & \textbf{SSIM} $\uparrow$ & \textbf{Mot. Fid.} $\uparrow$ & \textbf{RPE-T (norm.)} $\downarrow$ & \textbf{RPE-R ($^\circ$)} $\downarrow$ \\
\hline
Base LTX-2B (Zero-shot) & 391.14 & 0.520 & 12.61 & 0.37 & 0.40 & 0.036 & 1.39 \\
Real-Finetuned w/o AC-MoE       & 73.08  & 0.486 & \textbf{13.36} & \textbf{0.40} & 0.60 & 0.028 & 1.31 \\
Sim-Finetuned           & 72.72  & 0.514 & 13.03 & 0.39 & 0.67 & 0.051 & 1.73 \\
Real-Finetuned          & \textbf{65.39} & \textbf{0.481} & 13.27 & \textbf{0.40} & \textbf{0.72} & \textbf{0.024} & \textbf{1.18} \\

\hline
\multicolumn{8}{l}{\footnotesize \textit{Note:} Mot. Fid.: Motion Fidelity, RPE-T: Translational Relative Pose Error, \footnotesize RPE-R: Rotational Relative Pose Error.} \\
\multicolumn{8}{l}{ Best results are \textbf{bolded}.}
\end{tabular}
\vspace{-0.5cm}
\end{table*}

\subsection{Evaluation of Video Generation}

We assess the generalization capabilities of our video generation model using a real-world test set, distinct from the training distribution. This dataset encompasses diverse environmental scenes and varying lighting conditions, constituting a rigorous out-of-distribution evaluation.

The evaluation methodology mirrors the setup described in Section~\ref{sec:exp}. Our test corpus comprises 710 video clips, each containing 121 frames, extracted from approximately 12.3 minutes of raw videos. We evaluate the synthesized videos through both comprehensive quantitative metrics and qualitative visual analysis.

\begin{figure*}[t]
    \centering
    
    % Scene A: 48% of the width to leave a small gap in the middle
    \begin{subfigure}[b]{0.49\textwidth}
        \centering
        \includegraphics[width=\textwidth]{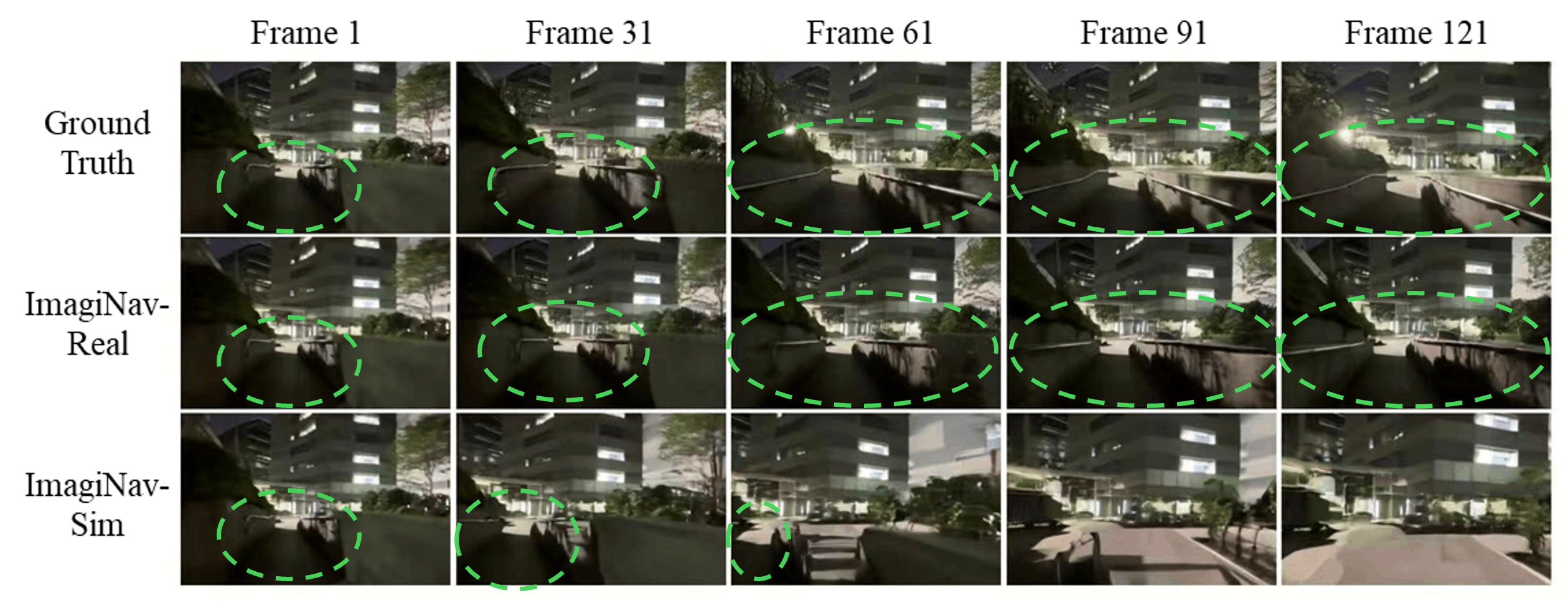}
        \caption{\textbf{Scene A: Navigable Affordance Perception.} Generation instruction: \textit{Move forward along the walkway towards the building.}}
        \label{fig:qual_cmp_guardrail}
    \end{subfigure}
    \hfill % This pushes the two subfigures to the edges
    % Scene B: 48% of the width
    \begin{subfigure}[b]{0.49\textwidth}
        \centering
        \includegraphics[width=\textwidth]{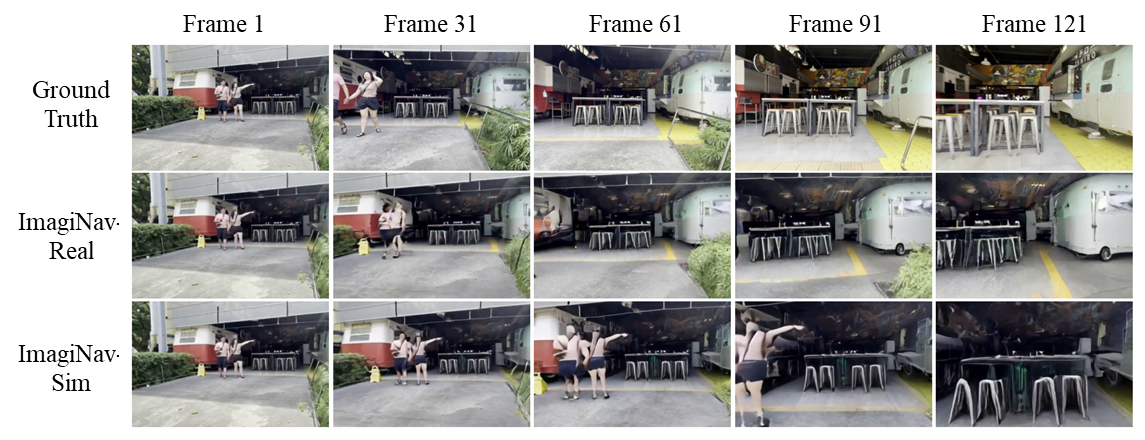}
        \caption{\textbf{Scene B: Dynamic Actor Modeling.} Generation instruction: \textit{Move forward and turn right towards the silver trailer.}}
        \label{fig:qual_cmp_pedestrian}
    \end{subfigure}
    
    \caption{\textbf{Qualitative Comparison of Simulation vs. Real-World Training.} We visualize the generated future trajectories across Ground Truth (Top Row), ImagiNav-Real (Middle Row), and ImagiNav-Sim (Bottom Row). Real-world data demonstrates superior understanding of static geometry (Scene A, the navigable walkway is circled in \textcolor{green}{green}), and dynamic behavior (Scene B).}
    \label{fig:qual_cmp}
    \vspace{-0.6cm}
\end{figure*}

\begin{figure}[t]
    \centering
    \includegraphics[width=\linewidth]{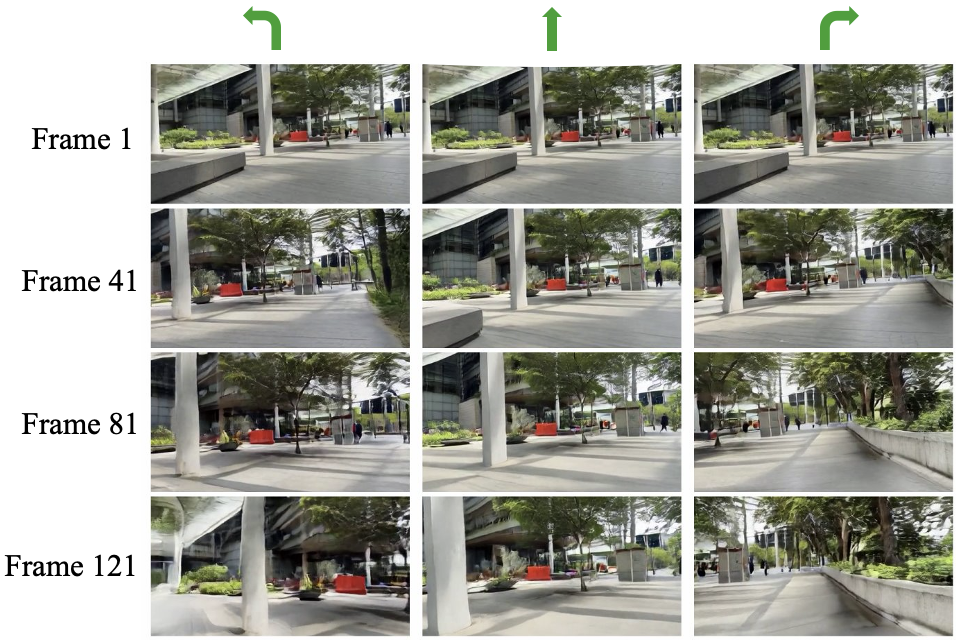}
    \caption{\textbf{Controllability and Geometric Grounding.} Given a single start frame, the model generates diverse trajectories conditioned on different language instructions (from left to right: \textit{``Move forward and turn left''}, \textit{``Move forward''}, and \textit{``Move forward and turn right''}).}
    \label{fig:qual_diversity}
    \vspace{-0.5cm}
\end{figure}

\subsubsection{Quantitative Results}

To quantify visual quality, we report standard video generation metrics: Fr\'echet Video Distance (FVD) \cite{ge2024content}, measuring distributional divergence of spatiotemporal dynamics, Learned Perceptual Image Patch Similarity (LPIPS) for perceptual consistency, and Peak Signal-to-Noise Ratio (PSNR) alongside Structural Similarity Index Measure (SSIM) for low-level reconstruction fidelity.

While the aforementioned metrics assess visual realism, they do not explicitly measure the kinematic accuracy required for downstream navigation tasks. For our visual planner, it is imperative that the synthesized video accurately reflects the intended motion dynamics rather than just visual plausibility. Consequently, we introduce a specific evaluation of kinematic fidelity based on the Relative Pose Error (RPE) and \textit{Motion Fidelity}. The ground-truth camera extrinsics are estimated via VGGT. Let $\mathbf{p}_{ref}^{(t)} = (x_{ref}^{(t)}, y_{ref}^{(t)}, \theta_{ref}^{(t)})$ and $\mathbf{p}_{gen}^{(t)} = (x_{gen}^{(t)}, y_{gen}^{(t)}, \theta_{gen}^{(t)})$ denote the reference and generated poses at time step $t$, respectively. We define the Translational RPE ($\text{RPE}_{trans}$) and Rotational RPE ($\text{RPE}_{rot}$) as the mean error over the trajectory of length $N$:

\vspace{-0.5cm}
\begin{equation}
    \text{RPE}_{trans} = \frac{1}{N} \sum_{t=1}^{N} \sqrt{(x_{ref}^{(t)} - x_{gen}^{(t)})^2 + (y_{ref}^{(t)} - y_{gen}^{(t)})^2},
\end{equation}
\vspace{-0.2cm}
\begin{equation}
    \text{RPE}_{rot} = \frac{1}{N} \sum_{t=1}^{N} \left| \text{wrap}(\theta_{ref}^{(t)} - \theta_{gen}^{(t)}) \right|,
\end{equation}
where $\text{wrap}(\cdot)$ normalizes the angular difference to the range $[-180^\circ, 180^\circ]$. $\text{RPE}_{trans}$ measures the Euclidean distance error in VGGT-normalized units, while $\text{RPE}_{rot}$ captures the absolute yaw deviation in degrees.

While RPE captures precise geometric deviations, it is sensitive to domain-specific kinematic shifts (e.g., velocity mismatches between simulation and reality), often penalizing semantically correct actions. To evaluate high-level behavioral alignment, we introduce Motion Fidelity, in which we discretize the continuous relative trajectories into a sequence of semantic action primitives $L = \{a_t\}_{t=1}^N$. Each action $a_t$ is assigned to one of four categories based on empirically determined thresholds: Forward (\texttt{F}), Turn Left (\texttt{TL}), Turn Right (\texttt{TR}), or Static (\texttt{S}). Motion Fidelity quantifies the alignment between the reference sequence $L_{\text{ref}}$ and the generated sequence $L_{\text{gen}}$ using the normalized Levenshtein distance:

\vspace{-0.5cm}
\begin{equation}
\text{Motion Fidelity} = 1 - \frac{\text{Levenshtein}(L_{\text{ref}}, L_{\text{gen}})}{\max(|L_{\text{ref}}|, |L_{\text{gen}}|)},
\end{equation}
where $\text{Levenshtein}(\cdot)$ computes the minimum edit distance. Because $L_{\text{ref}}$ and $L_{\text{gen}}$ are of equal length in our formulation, this distance simplifies to point-wise semantic substitutions.

Quantitative results (Table~\ref{tab:video_metrics}) reveal a distinct performance hierarchy. Because our objective is reliable visual planning rather than mere frame-level synthesis, we prioritize spatiotemporal consistency (FVD) and behavioral alignment (Motion Fidelity) as the primary indicators of success. The Base LTX-2B exhibits poor performance across all metrics, confirming that generic video models lack the fine-grained motion priors required for navigation. Our AC-MoE strategy efficiently establishes these priors, outperforming a single-model baseline in FVD and Motion Fidelity. While Sim-Finetuning improves on \emph{frame-level visual metrics} (PSNR, SSIM, LPIPS), it lags significantly in \emph{spatiotemporal consistency} (FVD 72.72 vs. 65.39) and \emph{kinematic realism} (Motion Fidelity 0.67 vs. 0.72). This gap reflects the dual limitations of simulation: the inherent lack of environmental diversity and the presence of synthetic kinematic artifacts. In contrast, the Real-Finetuned model dominates across all categories. By leveraging diverse human data, it captures natural kinematic flow and generalizes effectively to unseen environments.

\subsubsection{Qualitative Results}

We provide a qualitative comparison between models trained on simulation versus real-world data in Figure~\ref{fig:qual_cmp}. As shown in the bottom row (Scene A), the ImagiNav-Sim model exhibits brittle behavior in unstructured environments, hallucinating unsafe trajectories that collide with the guardrail. In contrast, ImagiNav-Real (middle row) accurately perceives the walkable affordance, generating a safe path that strictly adheres to the walkway.

Furthermore, Scene B highlights the critical role of real-world data in modeling dynamic scenes. While the simulation-trained model (bottom row) treats pedestrians as static objects due to the lack of dynamic actors in its training distribution, ImagiNav-Real (middle row) successfully anticipates human motion, generating physically consistent walking trajectories for the pedestrians in the scene.

Figure~\ref{fig:qual_diversity} demonstrates the controllability and geometric grounding of our real-finetuned generation model. Given an initial frame, the model synthesizes diverse future trajectories conditioned on distinct language instructions (\textit{``Move forward''}, \textit{``Move forward and turn left/right''}). Notably, the generation is not merely a rigid execution of motion primitives; it adapts to the environment. For instance, under the ``Move forward and turn left'' instruction, the model first steers slightly right to navigate around a stone bench before executing the turn. This demonstrates that the model does not blindly follow the text, but rather grounds the instruction in the physical geometry of the scene for collision avoidance.

\section{CONCLUSION}

In this work, we introduced \textbf{ImagiNav}, a hierarchical framework that redefines VLN by treating video generation as an embodiment-agnostic planner. By formulating navigation as a video prediction task, our approach leverages the rich semantic priors inherent in large-scale video models while enabling a scalable data pipeline that transforms readily available in-the-wild videos into actionable navigation policies. Our analysis demonstrates that the superior environmental diversity and kinematic naturalness of in-the-wild human data enable robust zero-shot generalization to physical robot navigation, paving the way for generalist robots that learn directly from human visual experience.

\textbf{Limitations and Future Work.} The primary limitation of our pipeline is inference latency, where the high computational cost of video generation currently restricts the system to a low-frequency control regime, limiting reactivity to sudden dynamic hazards. Additionally, operating purely in RGB space can lead to geometric hallucinations, where the model synthesizes visually plausible but physically unsafe trajectories; these may be mitigated through a visual plan scorer or traditional safety fallbacks. Finally, while our AC-MoE strategy successfully mitigates spatial ambiguity, a notoriously difficult and persistent challenge for generative video models, it remains a preliminary solution that introduces additional overhead by requiring the management of multiple experts. Future work will focus on: (1) distilling the heavy generative model into a lightweight, real-time policy to enable high-frequency closed-loop control; (2) integrating explicit depth modalities into the generation process to enforce stricter geometric consistency and safety in complex 3D environments; and (3) overcoming the fundamental limits of spatial reasoning by distillation of specialized experts into a single, unified model that maintains directional accuracy.

%%%%%%%%%%%%%%%%%%%%%%%%%%%%%%%%%%%%%%%%%%%%%%%%%%%%%%%%%%%%%%%%%%%%%%%%%%%%%%%%
%%%%%%%%%%%%%%%%%%%%%%%%%%%%%%%%%%%%%%%%%%%%%%%%%%%%%%%%%%%%%%%%%%%%%%%%%%%%%%%%

\section*{ACKNOWLEDGMENT}

This research is partly supported by the National Research Foundation, Singapore, under its Thematic CRP Programme, Award No. NRF-T-CRP-2025-0008. The authors acknowledge the use of AI tools (Gemini) for linguistic refinement and assistance with formatting. All suggestions were refined and verified by the authors, who take full responsibility for the paper’s content, data, and conclusions.
% \vspace{-0.2cm}

% We used ChatGPT to assist in the linguistic refinement of the Introduction and Conclusion sections. The generated text was critically reviewed and revised by the authors to ensure alignment with the research findings and academic standards.

%%%%%%%%%%%%%%%%%%%%%%%%%%%%%%%%%%%%%%%%%%%%%%%%%%%%%%%%%%%%%%%%%%%%%%%%%%%%%%%%
%%%%%%%%%%%%%%%%%%%%%%%%%%%%%%%%%%%%%%%%%%%%%%%%%%%%%%%%%%%%%%%%%%%%%%%%%%%%%%%%

\bibliographystyle{IEEEtran}

% This points to your .bib file (do not add the .bib extension here)
\bibliography{references}

\end{document}